\documentstyle[nips]{article}

\title{On-line Policy Improvement using Monte-Carlo Search}

\author{Gerald Tesauro \\
IBM T. J. Watson Research Center\\
P. O. Box 704\\
Yorktown Heights, NY  10598\\
tesauro@watson.ibm.com
\And
Gregory R. Galperin\\
MIT AI Lab\\
545 Technology Square\\
Cambridge, MA  02139\\
grg@ai.mit.edu
}

\begin{document}

\maketitle

\begin{abstract}
We present a Monte-Carlo simulation algorithm for real-time policy
improvement of an adaptive controller.  In the Monte-Carlo
simulation, the long-term expected reward of each possible action
is statistically measured, using the initial policy to make
decisions in each step of the simulation.  The action maximizing the
measured expected reward is then taken, resulting in an improved
policy.  Our algorithm is easily parallelizable
and has been implemented on the IBM SP1 and SP2 parallel-RISC
supercomputers.

We have obtained promising initial results in applying this
algorithm to the domain of backgammon.  Results are reported
for a wide variety of initial policies,
ranging from a random policy to TD-Gammon,
an extremely strong multi-layer neural network.  In each case,
the Monte-Carlo algorithm gives a substantial reduction, by
as much as a factor of 5 or more, in the error rate of
the base players.  The algorithm is also potentially useful
in many other adaptive control applications in which it
is possible to simulate the environment.
\end{abstract}

\section{INTRODUCTION}
Policy iteration, a widely used algorithm for solving problems in
adaptive control, consists of repeatedly iterating the following
policy improvement computation (Bertsekas, 1987):
(1)~First, a value function is computed that represents the
long-term expected reward that would be obtained by following
an initial policy.  (This may be done in several ways,
such as with the standard dynamic programming algorithm.)  (2)~An
improved policy is then defined which is greedy with respect to that
value function.  Policy iteration is known to have rapid and robust
convergence properties, and for Markov tasks with lookup-table
state-space representations, it is guaranteed to convergence to the
optimal policy.

In typical uses of policy iteration, the policy improvement
step is an extensive off-line procedure.  For example, in
dynamic programming, one performs a sweep through all states
in the state space.  Reinforcement learning provides another
approach to policy improvement; recently, several authors
have investigated using RL in conjunction with nonlinear
function approximators to represent the value functions and/or policies
(Tesauro, 1992; Crites and Barto, 1996; Zhang and Dietterich, 1996).
These studies are based on following
actual state-space trajectories rather
than sweeps through the full state space, but are still
too slow to compute improved policies in real time.
Such function approximators typically need extensive off-line
training on many trajectories before they achieve acceptable
performance levels.

In contrast, we propose an on-line algorithm for computing an improved
policy in real time.  We use Monte-Carlo search to estimate $V_P
(x,a)$, the expected value of performing action $a$ in state $x$ and
subsequently executing policy $P$ in all successor states.  Here, $P$
is some given arbitrary policy, as defined by a ``base controller''
(we do not care how $P$ is defined or was derived; we only need access
to its policy decisions).  In the Monte-Carlo search, many simulated
trajectories starting from $(x,a)$ are generated following $P$, and
the expected long-term reward is estimated by averaging the results
from each of the trajectories.  (Note that Monte-Carlo sampling is
needed only for non-deterministic tasks, because in a deterministic
task, only one trajectory starting from $(x,a)$ would need to be
examined.)  Having estimated $V_P (x,a)$, the improved policy $P'$ at
state $x$ is defined to be the action which produced the best
estimated value in the Monte-Carlo simulation, i.e., $P' (x) =
\arg\max_{a} V_P (x,a)$.

\subsection{EFFICIENT IMPLEMENTATION}
The proposed Monte-Carlo algorithm could be very
CPU-intensive, depending on the number of initial
actions that need to be simulated, the number of time
steps per trial needed to obtain a meaningful
long-term reward, the amount of CPU per time step
needed to make a decision with the base controller,
and the total number of trials needed to make a Monte-Carlo
decision.  The last factor depends on both the
variance in expected reward per trial, and on
how close the values of competing candidate actions are.

We propose two methods to address the potentially large
CPU requirements of this approach.  First, the power
of parallelism can be exploited very effectively.
The algorithm is easily parallelized with high efficiency:
the individual Monte-Carlo trials can be performed independently,
and the combining of results from different trials is
a simple averaging operation.  Hence there is relatively
little communication between processors required in a
parallel implementation.

The second technique is to continually monitor the
accumulated Monte-Carlo statistics during the simulation,
and to prune away both candidate actions that are sufficiently
unlikely (outside some user-specified confidence bound)
to be selected as the best action, as well as candidates whose
values are sufficiently close to the value of the current best
estimate that they are considered equivalent (i.e., choosing either
would not make a significant difference).  This technique
requires more communication in a parallel implementation,
but offers potentially large savings in the number of
trials needed to make a decision.

\section{APPLICATION TO BACKGAMMON}
We have initially applied the Monte-Carlo algorithm
to making move decisions in the game of backgammon.
This is an absorbing Markov process with perfect state-space
information, and one has a perfect model of the nondeterminism
in the system, as well as the mapping from actions to
resulting states.

In backgammon parlance, the expected value of a position is known as
the ``equity'' of the position, and estimating the equity by
Monte-Carlo sampling is known as performing a ``rollout.''  This
involves playing the position out to completion many times with
different random dice sequences, using a fixed policy $P$ to make move
decisions for both sides.  The sequences are terminated at the end of
the game (when one side has borne off all 15 checkers), and at that
time a signed outcome value (called ``points'') is recorded.  The
outcome value is positive if one side wins and negative if the other
side wins, and the magnitude of the value can be either 1, 2, or 3,
depending on whether the win was normal, a gammon, or a backgammon.
With normal human play, games typically last on the order of 50-60
time steps.  Hence if one is using the Monte-Carlo player to play out
actual games, the Monte-Carlo trials will on average start out
somewhere in the middle of a game, and take about 25-30 time steps to
reach completion.

In backgammon there are on average about 20 legal moves to consider in
a typical decision.  The candidate plays frequently differ in expected
value by on the order of .01. Thus in order to resolve the best play
by Monte-Carlo sampling, one would need on the order of 10K or more
trials per candidate, or a total of hundreds of thousands of
Monte-Carlo trials to make one move decision.  With extensive
statistical pruning as discussed previously, this can be reduced to
several tens of thousands of trials.  Multiplying this by 25-30
decisions per trial with the base player, we find that about a million
base-player decisions have to be made in order to make one Monte-Carlo
decision.  With typical human tournament players taking about 10
seconds per move, we need to parallelize to the point that we can
achieve at least 100K base-player decisions per second.

Our Monte-Carlo simulations were performed on the IBM
SP1 and SP2 parallel-RISC supercomputers at IBM Watson
and at Argonne National Laboratories.  Each SP node
is equivalent to a fast RS/6000, with floating-point
capability on the order of 100 Mflops.  Typical runs
were on configurations of 16-32 SP nodes, with parallel
speedup efficiencies on the order of 90\%.

We have used a variety of base players in our Monte-Carlo
simulations, with widely varying playing abilities and
CPU requirements.  The weakest (and fastest) of these
is a purely random player.  We have also used a few
single-layer networks (i.e., no hidden units) with
simple encodings of the board state, that were trained
by back-propagation on an expert data set (Tesauro, 1989).
These simple networks also make fast move decisions, and are much
stronger than a random player, but in human terms are
only at a beginner-to-intermediate level.  Finally, we
used some multi-layer nets with a rich input
representation, encoding both the raw board state and
many hand-crafted features, trained on self-play using
the TD($\lambda$) algorithm (Sutton, 1988; Tesauro, 1992).
Such networks play at an
advanced level, but are too slow to make Monte-Carlo
decisions in real time based on full rollouts to completion.
Results for all these players are presented in
the following two sections.

\subsection{RESULTS FOR SINGLE-LAYER NETWORKS}

We measured the game-playing strength of three single-layer base
players, and of the corresponding Monte-Carlo players, by playing
several thousand games against a common benchmark opponent.  The
benchmark opponent was TD-Gammon 2.1 (Tesauro, 1995), playing on its
most basic playing level (1-ply search, i.e., no lookahead).  Table
\ref{tbl-lin-ppg} shows the results.  Lin-1 is a single-layer neural
net with only the raw board description (number of White and Black
checkers at each location) as input.  Lin-2 uses the same network
structure and weights as Lin-1, plus a significant amount of random
noise was added to the evaluation function, in order to deliberately
weaken its playing ability.  These networks were highly optimized for
speed, and are capable of making a move decision in about 0.2 msec on
a single SP1 node.  Lin-3 uses the same raw board input as the other
two players, plus it has a few additional hand-crafted features
related to the probability of a checker being hit; there is no noise
added.  This network is a significantly stronger player, but is about
twice as slow in making move decisions.

\begin{table}
\begin{center}
\begin{tabular}{|c|c|c|c|} \hline
Network  & Base player & Monte-Carlo player & Monte-Carlo CPU \\ \hline
Lin-1 & -0.52 ppg & -0.01 ppg & 5 sec/move \\
Lin-2 & -0.65 ppg & -0.02 ppg & 5 sec/move \\
Lin-3 & -0.32 ppg & +0.04 ppg & 10 sec/move \\ \hline
\end{tabular}
\caption{ \label{tbl-lin-ppg}
Performance
of three simple linear evaluators,
for both initial base players and corresponding Monte-Carlo
players.  Performance is measured in terms of expected points
per game (ppg) vs. TD-Gammon 2.1 1-ply.  Positive numbers indicate
that the player here is better than TD-Gammon.  Base player stats
are the results of 30K trials (std. dev. about .005), and
Monte-Carlo stats are the results of 5K trials (std. dev.
about .02).  CPU times are for the Monte-Carlo player
running on 32 SP1 nodes.
}
\end{center}
\end{table}

We can see in Table \ref{tbl-lin-ppg}
that the Monte-Carlo technique produces
dramatic improvement in playing ability for these weak
initial players.  As base players, Lin-1 should be
regarded as a bad intermediate player, while Lin-2 is
substantially worse and is probably about equal to a human
beginner.  Both of these networks get trounced by TD-Gammon,
which on its 1-ply level plays at strong advanced level.
Yet the resulting Monte-Carlo players from these networks
appear to play about equal to TD-Gammon 1-ply.  Lin-3
is a significantly stronger player, and the resulting
Monte-Carlo player appears to be clearly better than TD-Gammon 1-ply.
It is estimated to be about equivalent to TD-Gammon on its 2-ply level,
which plays at a strong expert level.

The Monte-Carlo benchmarks reported in Table
\ref{tbl-lin-ppg} involved substantial amounts of CPU
time.  At 10 seconds per move decision, and 25 move
decisions per game, playing 5000 games against TD-Gammon
required about 350 hours of 32-node SP machine time.
We have also developed an alternative testing procedure,
which is much less expensive in CPU time, but still seems
to give a reasonably accurate measure of performance
strength.  We measure the average equity loss
of the Monte-Carlo player on a suite of test positions.
We have a collection of about 800 test positions, in
which every legal play has been extensively rolled out
by TD-Gammon 2.1 1-ply.  We then use the TD-Gammon rollout data to
grade the quality of a given player's move decisions.

\begin{table}
\begin{center}
\begin{tabular}{|c|c|c|c|} \hline
Evaluator  & Base loss & Monte-Carlo loss & Ratio \\ \hline
Random & 0.330 & 0.131 & 2.5  \\
Lin-1 & 0.040 & 0.0124 & 3.2 \\
Lin-2 & 0.0665 & 0.0175 & 3.8  \\
Lin-3 & 0.0291 & 0.00749 & 3.9 \\ \hline
\end{tabular}
\caption{ \label{tbl-lin-testset}
Average equity loss per move decision on an 800-position test set,
for both initial base players and corresponding Monte-Carlo
players.  Units are ppg; smaller loss values are better.
Also computed is ratio of base player loss to
Monte-Carlo loss.
}
\end{center}
\end{table}

Test set results for the three linear evaluators, and for
a random evaluator, are displayed in Table \ref{tbl-lin-testset}.
It is interesting to note for comparison that the TD-Gammon
1-ply base player scores 0.0120 on this test set measure, comparable
to the Lin-1 Monte-Carlo player, while TD-Gammon 2-ply base player scores
0.00843, comparable to the Lin-3 Monte-Carlo player.
These results are exactly in line with what we measured
in Table \ref{tbl-lin-ppg} using full-game benchmarking,
and thus indicate that the test-set methodology is in
fact reasonably accurate.  We also note that in each
case, there is a huge error reduction of potentially
a factor of 4 or more in using the Monte-Carlo technique.
In fact, the rollouts summarized in Table \ref{tbl-lin-testset}
were done using fairly aggressive statistical pruning;
we expect that rolling out decisions more extensively
would give error reduction ratios closer to factor of 5,
albeit at a cost of increased CPU time.

\subsection{RESULTS FOR MULTI-LAYER NETWORKS}

Using large multi-layer networks to do full rollouts is not feasible
for real-time move decisions, since the large networks are at least a
factor of 100 slower than the linear evaluators described previously.
We have therefore investigated an alternative Monte-Carlo algorithm,
using so-called ``truncated rollouts.''  In this technique trials are
not played out to completion, but instead only a few steps in the
simulation are taken, and the neural net's equity estimate of the
final position reached is used instead of the actual outcome.  The
truncated rollout algorithm requires much less CPU time, due to two
factors: First, there are potentially many fewer steps per trial.
Second, there is much less variance per trial, since only a few random
steps are taken and a real-valued estimate is recorded, rather than
many random steps and an integer final outcome.  These two factors
combine to give at least an order of magnitude speed-up compared to
full rollouts, while still giving a large error reduction relative to
the base player.

Table \ref{tbl-hid-testset} shows truncated rollout results
for two multi-layer networks: TD-Gammon 2.1 1-ply, which has 80 hidden
units, and a substantially smaller network with the same
input features but only 10 hidden units.  The first line of
data for each network reflects very extensive rollouts and shows
quite large error reduction ratios, although the CPU times
are somewhat slower than acceptable for real-time play.
(Also we should be somewhat suspicious of the 80 hidden unit
result, since this was the same network that generated the
data being used to grade the Monte-Carlo players.)
The second line of data shows what happens when the
rollout trials are cut off more aggressively.  This yields
significantly faster run-times, at the price of only
slightly worse move decisions.

\begin{table}
\begin{center}
\begin{tabular}{|c|c|l|c|c|} \hline
Hidden Units  & Base loss & Truncated Monte-Carlo loss & Ratio & M-C CPU \\ \hline
10 & 0.0152 & 0.00318 (11-step, thorough) & 4.8 & 25 sec/move\\
  &   & 0.00433 (11-step, optimistic) & 3.5 & 9 sec/move\\ \hline
80 & 0.0120 & 0.00181 (7-step, thorough) & 6.6 & 65 sec/move\\
  &   & 0.00269 (7-step, optimistic) & 4.5 & 18 sec/move\\ \hline
\end{tabular}
\caption{ \label{tbl-hid-testset}
Truncated rollout results for two multi-layer networks,
with number of hidden units and rollout steps as indicated.
Average equity loss per move decision on an 800-position test set,
for both initial base players and corresponding Monte-Carlo players.
Again, units are ppg, and smaller loss values are better.
Also computed is ratio of base player loss to
Monte-Carlo loss.  CPU times are for the Monte-Carlo player
running on 32 SP1 nodes.
}
\end{center}
\end{table}

The quality of play of the truncated rollout players shown in
Table \ref{tbl-hid-testset} is substantially better than
TD-Gammon 1-ply or 2-ply, and it is also substantially
better than the full-rollout Monte-Carlo players described
in the previous section.  In fact, we estimate that the
world's best human players would score in the range of 0.005 to 0.006
on this test set, so the truncated rollout players may actually
be exhibiting superhuman playing ability, in reasonable amounts
of SP machine time.

\section{DISCUSSION}

On-line search may provide a useful methodology for overcoming some of
the limitations of training nonlinear function approximators on
difficult control tasks.  The idea of using search to improve in real
time the performance of a heuristic controller is an old one, going
back at least to (Shannon, 1950).  Full-width search algorithms have
been extensively studied since the time of Shannon, and have produced
tremendous success in computer games such as chess, checkers and
Othello.  Their main drawback is that the required CPU time increases
exponentially with the depth of the search, i.e., $T \sim B^D$, where
$B$ is the effective branching factor and $D$ is the search depth.  In
contrast, Monte-Carlo search provides a tractable alternative for
doing very deep searches, since the CPU time 
for a full Monte-Carlo decision only scales as $T \sim
N\!\cdot\!B\cdot\!D$, where $N$ is the number of trials in the simulation.



In the backgammon application,
for a wide range of initial policies, our on-line Monte-Carlo
algorithm, which basically implements a single step of
policy iteration, was found to give very substantial error reductions.
Potentially 80\% or more of the base player's equity loss can be eliminated,
depending on how extensive the Monte-Carlo trials are.
The magnitude of the observed improvement is surprising to us:
while it is known theoretically that each step of policy iteration
produces a strict improvement, there are no guarantees on
how much improvement one can expect.  We have also noted
a rough trend in the data: as one increases the strength of
the base player, the ratio of error reduction
due to the Monte-Carlo technique appears to increase.
This could reflect superlinear convergence properties of
policy iteration.

In cases where the base player employs an evaluator that is
able to estimate expected outcome, the truncated rollout
algorithm appears to offer favorable tradeoffs relative to doing
full rollouts to completion.  While the quality of
Monte-Carlo decisions is not as good using truncated rollouts
(presumably because the neural net's estimates are biased),
the degradation in quality is fairly small in at least some cases,
and is compensated by a great reduction in CPU time.
This allows more sophisticated
(and thus slower) base players to be used, resulting in decisions
which appear to be both better and faster.

The Monte-Carlo backgammon program as implemented on the SP
offers the potential to achieve real-time move decision
performance that exceeds human capabilities.  In future work, we plan
to augment the program with a similar Monte-Carlo algorithm
for making doubling decisions.  It is quite possible that
such a program would be by far the world's best
backgammon player.

Beyond the backgammon application, we conjecture that
on-line Monte-Carlo search may prove to be useful in many other
applications of reinforcement learning and adaptive control.
The main requirement is that it should be possible to
simulate the environment in which the controller operates.
Since basically all of the recent successful applications of
reinforcement learning have been based on training in simulators,
this doesn't seem to be an undue burden.  Thus, for
example, Monte-Carlo search may well improve decision-making
in the domains of elevator dispatch (Crites and Barto, 1996)
and job-shop scheduling (Zhang and Dietterich, 1996).

We are additionally investigating two techniques for training a
controller based on the Monte-Carlo estimates.  First, one could train
each candidate position on its computed rollout equity, yielding a
procedure similar in spirit to TD($1$).  We expect this to converge to
the same policy as other TD($\lambda$) approaches, perhaps more
efficiently due to the decreased variance in the target values as well
as the easily parallelizable nature of the algorithm.  Alternately,
the base position -- the initial position from which the candidate
moves are being made -- could be trained with the best equity value
from among all the candidates (corresponding to the move chosen by the
rollout player).  In contrast, TD($\lambda$) effectively trains the
base position with the equity of the move chosen by the base
controller.  Because the improved choice of move achieved by the rollout
player yields an expectation closer to the true (optimal) value, we expect
the learned policy to differ from, and possibly be closer to optimal
than, the original policy.

\subsubsection*{Acknowledgments}
We thank Argonne National Laboratories for providing SP1 machine
time used to perform some of the experiments reported here.  Gregory
Galperin acknowledges support under Navy-ONR grant N00014-96-1-0311.

\subsubsection*{References}

D. Bertsekas, Dynamic Programming: Deterministic and Stochastic
Models.  Prentice-Hall (1987).

R. H. Crites and A. G. Barto, ``Improving elevator performance
using reinforcement learning.''  To appear in:
D. Touretzky et al., eds., Advances in Neural Information
Processing Systems 8, MIT Press (1996).

C. E. Shannon, ``Programming a computer for playing chess.''
{\em Philosophical Magazine} {\bf 41}, 265-275 (1950).

R. S. Sutton, ``Learning to predict by the methods of
temporal differences.''  {\em Machine Learning} {\bf 3}, 9-44 (1988).

G. Tesauro, ``Connectionist learning of expert preferences
by comparison training.'' In: D. Touretzky, ed., Advances in
Neural Information Processing Systems 1, 99-106,
Morgan Kaufmann (1989).

G. Tesauro, ``Practical issues in temporal difference learning.''
{\em Machine Learning} {\bf 8}, 257-277 (1992).

G. Tesauro, ``Temporal difference learning and TD-Gammon.''
{\em Comm. of the ACM}, {\bf 38:3}, 58-67 (1995).

W. Zhang and T. G. Dietterich, ``High-performance job-shop
scheduling with a time-delay TD($\lambda$) network.''  To appear in:
D. Touretzky et al., eds., Advances in Neural Information
Processing Systems 8, MIT Press (1996).

\end{document}